\documentclass[preprint]{article}

\usepackage{microtype}
\usepackage{graphicx}
\usepackage{subcaption}
\usepackage{booktabs}
\usepackage{xcolor}

\usepackage{hyperref}

\usepackage{icml2026}

\usepackage{amsmath}
\usepackage{amssymb}
\usepackage{mathtools}
\usepackage{amsthm}

\usepackage{tcolorbox}
\usepackage{multirow}
\usepackage{array}

\usepackage[capitalize,noabbrev]{cleveref}

\theoremstyle{plain}

\theoremstyle{definition}

\theoremstyle{remark}

\icmltitlerunning{Three-Axis Fidelity for Aligning LLM-Based Survey Simulators from Small Pilot Data}

\begin{document}

\twocolumn[
  \icmltitle{Beyond the Mean: Three-Axis Fidelity for \\Aligning LLM-Based Survey Simulators from Small Pilot Data}

  \icmlsetsymbol{equal}{*}

  \begin{icmlauthorlist}
    \icmlauthor{Eun Cheol Choi}{usc}
    \icmlauthor{Youngrae Kim}{usc}
    \icmlauthor{Prabhu Pugalenthi}{usc}
    \icmlauthor{Hong-En Chen}{usc}
    \icmlauthor{Bo-Ruei Huang}{usc}
  \end{icmlauthorlist}

  \icmlaffiliation{usc}{University of Southern California, CA, USA}

  \icmlcorrespondingauthor{Eun Cheol Choi}{euncheol@usc.edu}

  \icmlkeywords{Large Language Models, Social Survey Simulation, Misinformation, LoRA, Concordance Correlation Coefficient, Prediction-Powered Inference}

  \vskip 0.3in
]

\printAffiliationsAndNotice{All authors contributed equally to this work.}

\begin{abstract}
Large language models (LLMs) are increasingly used to simulate social survey responses, yet their outputs exhibit systematic biases: marginal distributions are skewed, response variance is poorly calibrated, and predictor--outcome relationships are attenuated. We ask a simple question: given a small pilot sample of human responses, can an LLM recover the statistical characteristics of a broader population? We decompose recovery along three axes: structural fidelity, marginal fidelity, and individual fidelity. Using a COVID-19 misinformation survey as a case study, we benchmark three families of approaches: prompting, rectification, and fine-tuning. The findings suggest that fine-tuning on small pilot samples offers a balanced approach for achieving multiple forms of fidelity, but the levels of such fidelity can vary across subsamples, potentially threatening pluralistic alignment.
\end{abstract}

\section{Introduction}

Large language models (LLMs) are now routinely used as proxies for human respondents -- ranging from \emph{silicon samples} of voters \cite{argyle2023out} to interview-grounded ``generative agents'' that simulate $1{,}000$ real Americans \cite{park2024generative}, to fine-tuned models that predict experiment-level outcomes \cite{kolluri2025finetuning} and prompted GPT-4 used to forecast effect sizes from social-science experiments \cite{hewitt2024predicting}. In parallel, social surveys remain the dominant instrument for measuring beliefs and attitudes, but recruiting representative respondents is expensive and slow, so most large surveys begin with a small \emph{pilot} sample to gauge feasibility \cite{van2001importance}. This raises a concrete computational question: \emph{can an LLM, given a small human pilot, recover the statistical structure of the full population it was drawn from?}

This question is especially important in the domain of \emph{misinformation belief}, where (i) populations differ in which false claims they have even \emph{encountered} \cite{lee2023role}; (ii) accuracy ratings depend on local information ecosystems that pretraining may not fully capture \cite{choi2026overstating}; and (iii) downstream uses (e.g.\ targeted interventions based on certain psychosocial features) rely on \emph{predictor--outcome} relations, and not just on marginal distributions. Recent audits show that LLM-generated survey responses (i) approximate marginals but flatten variance, with effect-size signs flipping in roughly a third of cases \cite{bisbee2024synthetic}; (ii) exhibit topic-specific ``machine bias'' that is socially inconsistent across topics \cite{boelaert2025machine}; (iii) homogenize and structurally distort minority groups \cite{li2025chatgpt, wang2025large}; (iv) are highly sensitive to seemingly innocuous prompt perturbations \cite{rupprecht2025prompt, tjuatja2024llms}; and (v) display large analytic flexibility \cite{cummins2025threat}. Together, these pathologies suggest that LLMs should not be treated as drop-in replacements for human respondents, but rather as conditional generative or estimation models whose outputs must be evaluated against ground-truth data.

We make four contributions. \textbf{First}, we reframe LLM-based survey simulation as a \emph{recoverability} task -- given a small pilot $D_p$ of human responses, how much of a held-out population's structure can be reconstructed -- and decompose recovery along three axes: \emph{structural} fidelity (do predictor--outcome relations match?), \emph{marginal} fidelity (do simulated marginals match the human ones?), and \emph{individual} fidelity (does each simulated respondent track their human counterpart?). \textbf{Second}, we introduce a calibrated evaluation protocol that maps each fidelity axis to a specific metric: Lin's Concordance Correlation Coefficient (CCC), decomposed into a sign-agreement rate and a magnitude ratio, for the structural axis; cross-respondent Earth Mover's Distance (EMD) on per-respondent scalar summaries for the marginal axis; and paired Pearson $r_d$ (relative) and MAE$_d$ (absolute) on per-respondent discernment for the individual axis (Sec.~\ref{sec:eval}). \textbf{Third}, on the same 5\% pilot of an $N{=}1{,}466$ COVID-19 misinformation survey, we run a head-to-head comparison of $\{$ZS, FS$\}{\times}\{$batch, per-item$\}$ prompting, PPI rectification \cite{angelopoulos2023prediction} applied uniformly to every prompt-based simulator, and LoRA / LoRA $+$ MLP fine-tuning. \textbf{Fourth}, we audit fidelity \emph{within} each demographic subgroups: statistical recovery degrades for certain types of identity subgroups, calling for the attention of the pluralistic alignment community.

\section{Related Work}

We organize prior work into calibration families relevant to the small-pilot setting: prompt-based conditioning, test-time intervention and statistical rectification, and parameter-efficient fine-tuning.

\subsection{Prompt-Based Conditioning}

The earliest line of work conditions LLMs on textual descriptions of demographics, attitudes, or context. \emph{Silicon Sampling} \cite{argyle2023out} showed that GPT-3, prompted with ANES-style backstories, can approximate group-level voting distributions, and \emph{Random Silicon Sampling} \cite{sun2024random} extended this to demographic role-playing using group-level marginals. \emph{LLM-Mirror} \cite{kim2024llm} added pre-existing responses and psychological traits, and other richer schemes \cite{choi2026overstating} incorporate social-network and peer features. \citet{park2024generative} uses full two-hour interview transcripts as the persona and recovers $85\%$ of human test--retest accuracy on the GSS. \emph{Audience Segmentation} \cite{qin2026restoring} restores within-group heterogeneity by varying identifier granularity. The common assumption is that the LLM's \emph{Universal Prior} -- general world knowledge acquired during pretraining -- is rich enough that, conditioned on the right description, the model produces calibrated human-like responses.

A complementary literature audit tests that assumption and finds it fragile. \citet{bisbee2024synthetic} shows that ChatGPT-generated feeling thermometers compress variance and flip effect-size signs in $\sim$32\% of cases on ANES items; \citet{boelaert2025machine} documents opinion-poll ``machine bias'' that is socially inconsistent across topics; \citet{wang2025large} shows that identity-prompted LLMs harmfully misportray and flatten minority groups; \citet{li2025chatgpt} formalizes this as ``Das Man'' homogenization driven by accuracy-maximizing decoding. \citet{zhou2025chatgpt} report that, even with repeated random sampling from GPT, the resulting silicon population overrepresents some demographic groups and is far more deterministic than humans on attitudinal items. \citet{chapala2025mitigating} test prompt-based mitigations of social-desirability bias including neutral third-person reformulation and reverse-coding. \citet{cummins2025threat} stress-tests the entire pipeline, showing that 252 plausible analyst configurations yield strikingly different conclusions, motivating multiverse-style robustness checks.

A natural response to these audits is to inject a small amount of in-context examples drawn from the same population. This idea underlies recent few-shot demonstrations \cite{argyle2023out}, persona-pretest pipelines \cite{kim2024llm}, and audience-segmentation prompting \cite{qin2026restoring}.

\subsection{Test-Time Intervention and Rectification}

A second family of methods accepts that prompting alone is insufficient and instead modifies elicitation or post-processing. \emph{Semantic Similarity Rating} \cite{maier2025llms} avoids regression-to-the-mean on Likert scales by eliciting free text and projecting it into a similarity space. A statistically grounded sub-family treats LLM outputs as biased predictors and corrects them with a small gold sample. \emph{Prediction-Powered Inference} (PPI) \cite{angelopoulos2023prediction} debiases synthetic population estimates using a held-out human sample, and \citet{krsteski2025valid} adapts PPI for survey simulation with a power-tuned per-item $\lambda$. We use this PPI variant as our rectification family, applied uniformly to every prompt-based simulator.

\subsection{Fine-Tuning for Survey Simulation}

A third, rapidly growing family fine-tunes the LLM directly on human responses. \citet{kolluri2025finetuning} fine-tune LLaMA3-8B and Qwen2.5-14B on $\sim$2.9M responses from over 400{,}000 participants in the SocSci210 corpus, reducing prediction error on unseen experiments by 30\% and 26\% respectively relative to GPT-4o. \citet{cao2025specializing} fine-tune LLMs to match country-level WVS / Pew distributions using a first-token-probability objective and generalize to unseen questions and countries. \citet{huang2025distribution} introduces \emph{Distribution Shift Alignment}, a two-stage scheme that explicitly aligns subgroup-conditional shifts and reduces required real-data volume by $53$--$69\%$. 

\citet{krsteski2025valid} provides the closest direct comparison of \emph{prompting, fine-tuning, and rectification} under limited data, showing that the methods are complementary rather than substitutes. We differ in three ways: (i) our domain is COVID-19 misinformation, with predictor--outcome structure rich enough to expose multivariate failures invisible in marginal accuracy alone \cite{choi2026overstating}; (ii) alongside the population's marginal fidelity, we report structural-fidelity metrics that penalizes deviations from the bivariate and regression coefficients; (iii) we report how much individual predictions are correlated to each of the human participants' responses, providing a benchmark of individual fidelity as well. We also address whether the choice of \emph{output head} -- autoregressive token generation vs.\ a discriminative classification head -- matter for the simulation fidelity.

\section{Methodology}\label{sec:methodology}

We frame the problem as a recoverability task. Given a full survey dataset $D$, we observe a small pilot $D_p \subset D$ and aim to generate a synthetic dataset $D_s$ over the held-out respondents $D \setminus D_p$ such that $D_s$ preserves the statistical properties of $D$.

\subsection{Data}

The survey is the COVID-19 misinformation belief study of \citet{lee2023role}, conducted in South Korea in May 2020 ($N{=}1{,}466$). Each respondent $i$ is described by a profile $X_i$ comprising demographics (age, gender, education, income, political orientation), psychometric scales (open-mindedness, faith in intuition, need for evidence, truth-as-political, skepticism), and exposure measures (info exposure, emotional response). Each respondent answers $Y_i$ on 36 belief items: 18 \textsc{Misinfo} (false claims) and 18 \textsc{Trueinfo} (true claims), with 9 political and 9 scientific items in each subset. Responses use a 4-point Likert scale (\textit{Not accurate at all}, \textit{Not very accurate}, \textit{Somewhat accurate}, \textit{Very accurate}) plus a separate \textit{Have not seen it} (\textsc{Hns}) option that we treat as missing throughout. The empirical \textsc{Hns} rate is $13.3\%$. We define three per-respondent scalar summaries -- the \textsc{Misinfo} mean $m_i = \overline{y_{i,\text{Mis}}}$, the \textsc{Trueinfo} mean $t_i = \overline{y_{i,\text{Tru}}}$, and a \textsc{Discernment} score $d_i = t_i - m_i$ -- and use them throughout the structural, marginal, and individual fidelity analyses.

\subsection{Pilot Sampling}

We draw a 5\% pilot $D_p$ ($n{=}74$) with a fixed seed, holding out the remaining $1{,}392$ respondents as the evaluation set. The same pilot is reused across all calibration pipelines so that differences in performance reflect differences in method rather than data.

\subsection{Calibration Pipelines}

We benchmark four families. All upstream simulators predict the same 36 Likert ratings on the same held-out respondents.

\paragraph{Family 1 -- Zero-Shot Persona (ZS, ZS-perItem).}
For each held-out respondent $i$, the LLM is given $X_i$ and asked to predict $Y_i$. \textbf{ZS} (batch) elicits all 36 ratings in a single prompt; \textbf{ZS-perItem} elicits them one item at a time. Comparing the two isolates the effect of cross-item conditioning independently of pilot examples, ablating one of the analytic-flexibility knobs flagged by \citet{cummins2025threat}. An example prompt is presented in App.~\ref{app:prompts}.

\paragraph{Family 2 -- Few-Shot Prompting (FS, FS-perItem).}
We randomly inject a subset (five rows) of $D_p$ as in-context examples. Together with Family 1 these form a $2{\times}2$ factorial of $\{$no pilot, with pilot$\}{\times}\{$batch, per-item$\}$.

\paragraph{Family 3 -- Parameter-Efficient Fine-Tuning (LoRA, LoRA $+$ MLP).}
We fine-tune Qwen3-8B \cite{yang2025qwen3} with LoRA \cite{hu2022lora} on the same 5\% pilot in two configurations. \textbf{LoRA} -- adapters on attention and MLP projections, autoregressively emitting the label string. \textbf{LoRA $+$ MLP} -- the same LoRA adapters plus a trained 5-way MLP classification head on the final hidden state, with cross-entropy over the four Likert classes plus a fifth class for \textsc{Hns}. The two configurations differ only in output head.

\paragraph{Family 4 -- PPI Rectification.}
We rectify each prompt-based simulator's per-item population mean using Prediction-Powered Inference \cite{angelopoulos2023prediction}, following \citet{krsteski2025valid}'s adaptation to survey simulation. For each item $q$, the rectified per-item population estimate is
\[
\hat{\theta}_q = \bar{y}_{\text{pilot},q} + \lambda_q\,\bigl(\bar{\hat{y}}_{\text{held},q} - \bar{\hat{y}}_{\text{pilot},q}\bigr),
\]
with $\lambda_q = \mathrm{Cov}(y, \hat{y}) / \mathrm{Var}(\hat{y})$ on the pilot per item. PPI emits a scalar per item, not individual predictions, so it is evaluable only on per-subset population means (Tab.~\ref{tab:ppi}). For our LoRA configurations the simulator was trained on the pilot, so its in-distribution pilot predictions reduce to memorized GT and the correction term collapses; we report PPI and discuss the algebraic degeneracy in \S\ref{sec:pop}.

\subsection{Evaluation Metrics}\label{sec:eval}

We decompose recovery into three complementary axes. Each axis answers a different question, and a method that excels on one can fail on another. Details of the bootstrapping procedure and uncertainty intervals on every headline metric are reported in App.~\ref{app:full-cis}.

\paragraph{Structural fidelity.}
\emph{Do predictor--outcome relations match?} We regress $d_i$ on twelve predictors -- the seven psychometric / exposure scales plus five demographics (age, gender, education, income, political orientation) -- and assemble two paired vectors per method: the bivariate predictor--$d_i$ correlations, $\{(r_k^{\text{GT}}, r_k^{\text{sim}})\}_{k=1}^{12}$, and the standardized OLS coefficients. We calculate the GT--Sim Concordance Correlation Coefficient (CCC), which penalizes deviations from the y$=$x line and so jointly captures direction and magnitude:
\[
\rho_c = \frac{2\,\mathrm{Cov}(x,y)}{\mathrm{Var}(x)+\mathrm{Var}(y)+(\bar x-\bar y)^2}.
\]
Alongside CCC we report a sign-agreement rate (fraction of the 12 predictors with matching sign) and a magnitude ratio $\overline{|x^{\text{sim}}|}/\overline{|x^{\text{GT}}|}$ (\textgreater 1 inflates, \textless 1 compresses), giving a direction-vs-magnitude decomposition.

\paragraph{Marginal fidelity.}
\emph{Do simulated marginals match the human ones?} For each held-out respondent we take the three scalar summaries $d_i$, $m_i$, $t_i$ and report the Wasserstein-1 \emph{Earth Mover's Distance} (EMD) between the simulator's and GT's cross-respondent distributions of each summary. All three EMDs are distances between the same kind of object (cross-respondent distribution of a per-respondent scalar), so they are directly comparable. We supplement the EMDs with the per-subset population means $\mu_\text{Mis}$ and $\mu_\text{Tru}$ before and after PPI rectification (Tab.~\ref{tab:ppi}).

\paragraph{Individual fidelity.}
\emph{Does each simulated respondent track their human counterpart?} We summarize each respondent by the discernment scalar $d_i$ and report two paired statistics across respondents: a relative agreement metric $r_d = \mathrm{Pearson}(d_i^{\text{GT}}, d_i^{\text{sim}})$, which asks ``do high-discernment respondents get high-discernment predictions,'' and an absolute agreement metric $\text{MAE}_d = \overline{\,|d_i^{\text{GT}}-d_i^{\text{sim}}|\,}$ across respondents, which asks ``how far off is each respondent's predicted discernment.'' The two answer different questions: a simulator can rank respondents correctly while inflating magnitudes (high $r_d$, large MAE$_d$) or hit absolute values close on average without preserving the ranking (low MAE$_d$, low $r_d$). The same pair $(r, \text{MAE})$ is also reported on $m_i$ and $t_i$ in App.~\ref{app:full-cis}.

\section{Results}

\begin{table}[t]
\centering
\small
\setlength{\tabcolsep}{4pt}
\resizebox{\columnwidth}{!}{%
\begin{tabular}{l|cc|c|cc}
\toprule
 & \multicolumn{2}{c|}{\textbf{Structural}~$\uparrow$} & \textbf{Marginal}$\downarrow$ & \multicolumn{2}{c}{\textbf{Individual}}\\
\textbf{Method} & CCC$_{\text{biv}\,r}$ & CCC$_{\text{OLS}\,\beta}$ & EMD-$d$ & $r_d\,\uparrow$ & MAE$_d\,\downarrow$ \\
\midrule
ZS & 0.56 & 0.50 & 0.72 & 0.12 & 0.79 \\
ZS-perItem & 0.80 & 0.71 & 0.48 & 0.31 & 0.58 \\
FS & 0.51 & 0.59 & 0.23 & 0.18 & 0.67 \\
FS-perItem & 0.61 & 0.63 & 0.41 & 0.27 & \textbf{0.58} \\
LoRA & 0.60 & 0.41 & 0.63 & 0.17 & 0.75 \\
LoRA + MLP & \textbf{0.85} & \textbf{0.78} & \textbf{0.17} & \textbf{0.37} & 0.61 \\
\bottomrule
\end{tabular}}
\caption{\textbf{Headline metrics across the three fidelity axes} on the held-out evaluation set. \textbf{Structural}: Lin's CCC on $K{=}12$ paired predictor--$d_i$ correlations and 12 standardized OLS coefficients. \textbf{Marginal}: cross-respondent EMD on $d_i$. \textbf{Individual}: $r_d$ is the cross-respondent Pearson between simulator and GT $d_i$; MAE$_d$ is mean $|d_i^{\text{GT}}-d_i^{\text{sim}}|$. Bootstrapped CIs in App.~\ref{app:full-cis}.}
\label{tab:headline}
\end{table}

\subsection{Structural Fidelity}\label{sec:structural}

\begin{figure}[t]
\centering
\includegraphics[width=\columnwidth]{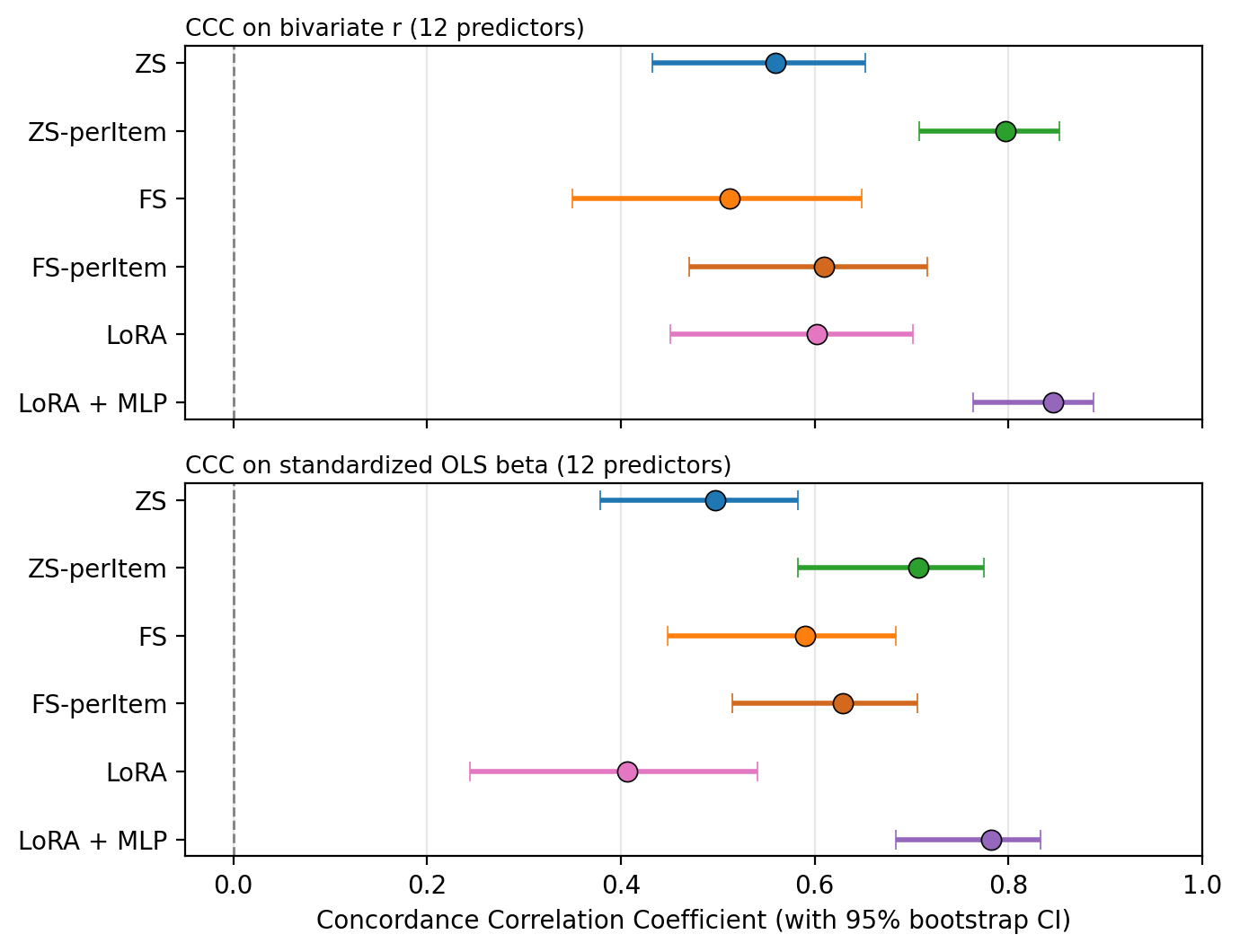}
\caption{\textbf{Structural-fidelity forest plot} on the 12 predictors. Per method: point estimate of Lin's CCC with bootstrapped CIs (higher = better; App.~\ref{app:full-cis}). Dashed line at $0$. Top panel = bivariate $r$; bottom panel = standardized OLS $\beta$.}
\label{fig:ccc-forest}
\end{figure}

Tab.~\ref{tab:headline} reports the headline numbers; Fig.~\ref{fig:ccc-forest} renders the structural axis as a forest plot. \textbf{LoRA $+$ MLP} shows the highest point estimates of bivariate-$r$ (CCC $0.85$) and the OLS-$\beta$ (CCC $0.78$). \textbf{ZS-perItem} is the strongest prompt-only competitor (CCC $0.80$ on bivariate $r$; $0.71$ on OLS $\beta$) -- with no pilot examples or fine-tuning.

\paragraph{Direction vs.\ magnitude (Tab.~\ref{tab:aux}).}
Decomposing the direction and the magnitude of relationships among variables reveals different success/failure modes hidden behind similar headline numbers. \textbf{LoRA $+$ MLP} and \textbf{FS-perItem} tie for the highest sign-agreement on bivariate $r$ ($0.92$, $11$ of $12$ predictors), but differ on magnitude: FS-perItem sits closest to the y$=$x line while LoRA $+$ MLP mildly inflates. At the other end, \textbf{LoRA} compresses slopes most severely, with both direction and magnitude off. The broader pattern is that high CCC requires \emph{both} directional agreement and matched magnitude, and that fine-tuning on the pilot appears to swing magnitudes in either direction (compression for LoRA, inflation for LoRA $+$ MLP). Some of the patterns are consistent with the structural failures (variance flattening, effect-size attenuation) documented in prior LLM-survey audits \cite{bisbee2024synthetic, choi2026overstating}. Per-method sign-agreement and magnitude ratios are in Tab.~\ref{tab:aux} and App.~\ref{app:full-cis} with bootstrapped CIs.

\begin{table}[t]
\centering
\small
\setlength{\tabcolsep}{6pt}
\resizebox{\columnwidth}{!}{%
\begin{tabular}{l|cc|cc}
\toprule
 & \multicolumn{2}{c|}{\textbf{Bivariate $r$}} & \multicolumn{2}{c}{\textbf{Standardized $\beta$}} \\
\textbf{Method} & sign-agree & $|$sim$|/|$GT$|$ & sign-agree & $|$sim$|/|$GT$|$ \\
\midrule
ZS & 0.58 & 1.07 & 0.58 & 1.03 \\
ZS-perItem & 0.83 & 1.48 & 0.75 & 1.50 \\
FS & 0.58 & 0.56 & 0.50 & 0.63 \\
FS-perItem & 0.92 & 0.85 & 0.67 & 0.99 \\
LoRA & 0.50 & 0.50 & 0.50 & 0.58 \\
LoRA + MLP & 0.92 & 1.28 & 0.75 & 1.32 \\
\bottomrule
\end{tabular}}
\caption{\textbf{Structural-fidelity decomposition.} Sign-agreement = fraction of the $K{=}12$ predictors with matching sign across simulator and GT (higher = better). $|$sim$|/|$GT$|$ is the ratio of mean absolute coefficients (\textgreater 1 inflates, \textless 1 compresses). Bootstrap CIs are in App.~\ref{app:full-cis}.}
\label{tab:aux}
\end{table}

\subsection{Marginal Fidelity}\label{sec:pop}

\begin{figure}[t]
\centering
\includegraphics[width=\columnwidth]{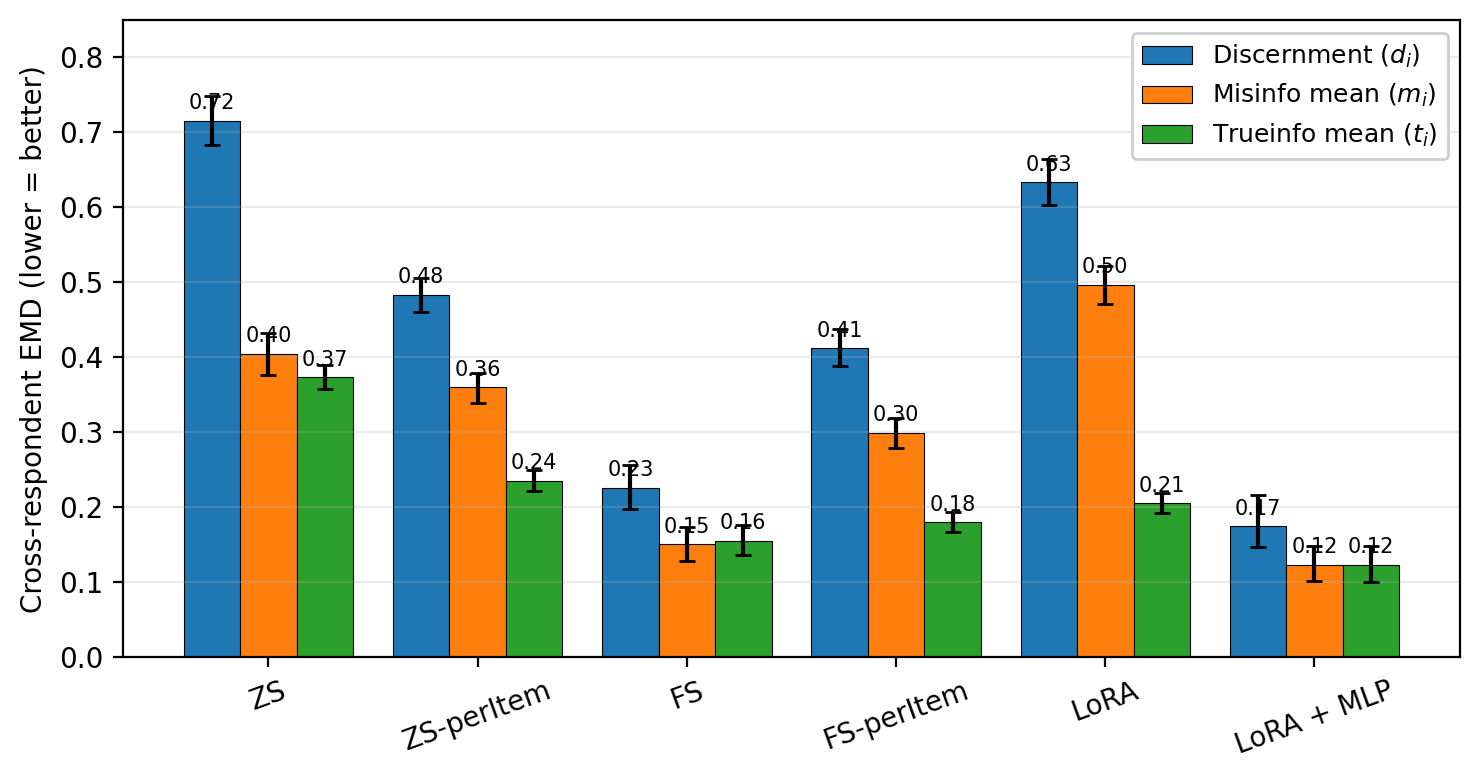}
\caption{\textbf{Cross-respondent EMD on per-respondent scalar summaries.} Wasserstein-1 between simulator and GT distributions of, respectively, discernment $d_i$, Misinfo mean $m_i$, and Trueinfo mean $t_i$. Lower is better; error bars are bootstrap uncertainty intervals (App.~\ref{app:full-cis}).}
\label{fig:emd-scalar}
\end{figure}

Fig.~\ref{fig:emd-scalar} plots the three cross-respondent EMDs. \textbf{LoRA $+$ MLP} achieves the lowest EMD on all three summaries, with \textbf{FS} second. Batch \textbf{ZS} is the weakest on discernment EMD, tending to move the two subset means in opposite directions (under-rating Misinfo, over-rating Trueinfo) and inflating the cross-respondent variance of $d_i$. Full per-summary values with CIs are in App.~\ref{app:full-cis}.

\paragraph{PPI rectification.}
Prediction-Powered Inference treats each upstream simulator's per-item population estimate as a biased predictor and rectifies it using the pilot, $\hat\theta_q = \bar{y}_{\text{pilot},q} + \lambda_q (\bar{\hat{y}}_{\text{held},q} - \bar{\hat{y}}_{\text{pilot},q})$, with $\lambda_q = \mathrm{Cov}(y, \hat y) / \mathrm{Var}(\hat y)$ on the pilot. In our run, PPI helps the most biased simulator (\textbf{ZS}, both subset means moving toward GT), is roughly neutral for already-calibrated methods, and can modestly hurt the Misinfo estimate of methods whose raw output is already near GT. For the LoRA family the rectifier is algebraically degenerate: when $\hat y_{\text{pilot}} = y_{\text{pilot}}$ (the limit of training-set memorization), $\lambda_q\equiv 1$ and $\hat\theta_q = \bar{\hat y}_{\text{held},q}$, so PPI reduces to the simulator's raw held-out mean; we mark these rows with $^\dagger$ in Tab.~\ref{tab:ppi}. The reading we draw is that rectification is a \emph{conditional} tool whose benefit depends on where the simulator's bias variance sits relative to the pilot's sampling variance. Per-item values with CIs are in App.~\ref{app:full-cis}.

\begin{table}[t]
\centering
\small
\setlength{\tabcolsep}{4pt}
\begin{tabular}{l|cc|cc}
\toprule
 & \multicolumn{2}{c|}{\textbf{Raw simulator}} & \multicolumn{2}{c}{\textbf{$+$ PPI rectification}}\\
\textbf{Method} & $\mu_\text{Mis}$ & $\mu_\text{Tru}$ & $\mu_\text{Mis}$ & $\mu_\text{Tru}$ \\
\midrule
\textit{GT} & 2.04 & 3.13 & 2.04 & 3.13 \\
\midrule
ZS & 1.77 & 3.45 & 2.12 \textcolor{green!50!black}{$\downarrow$} & 3.08 \textcolor{green!50!black}{$\downarrow$} \\
ZS-perItem & 2.01 & 2.98 & 2.12 \textcolor{red}{$\uparrow$} & 3.08 \textcolor{green!50!black}{$\downarrow$} \\
FS & 2.06 & 3.28 & 2.11 \textcolor{red}{$\uparrow$} & 3.09 \textcolor{green!50!black}{$\downarrow$} \\
FS-perItem & 2.02 & 3.13 & 2.11 \textcolor{red}{$\uparrow$} & 3.07 \textcolor{red}{$\uparrow$} \\
LoRA$^\dagger$ & 1.57 & 3.21 & 1.58  & 3.19  \\
LoRA + MLP$^\dagger$ & 2.01 & 3.23 & 2.03  & 3.21  \\
\bottomrule
\end{tabular}
\caption{\textbf{PPI rectification.} Per-subset population means: raw simulator output (left) and PPI-rectified estimates aggregated over the 18 items in each subset (right). $\downarrow$ marks subsets where PPI brings the estimate closer to GT than the raw simulator; $\uparrow$ marks the opposite. $^\dagger$ marks rows where the LoRA family was trained on the pilot, so the rectifier reduces algebraically to the raw held-out mean in the case of perfect memorization (see \S\ref{sec:pop}). }
\label{tab:ppi}
\end{table}

\subsection{Individual Fidelity}\label{sec:resp}

The individual-fidelity columns of Tab.~\ref{tab:headline} report two paired statistics on the held-out respondents' discernment scalars $d_i$. \textbf{LoRA $+$ MLP} wins the relative axis ($r_d{=}0.37$), with \textbf{ZS-perItem} second ($r_d{=}0.31$). The absolute axis is closer: \textbf{ZS-perItem} and \textbf{FS-perItem} attain the smallest per-respondent deviation ($\text{MAE}_d{=}0.58$), with \textbf{LoRA $+$ MLP} close behind. So ZS-perItem's predicted discernment is, on average, closest to GT in absolute terms, but its respondent ranking appears looser than LoRA $+$ MLP's. The full metrics are in App.~\ref{app:full-cis}.

\subsection{Subgroup Fidelity and the Pluralistic-Alignment Stake}\label{sec:subgroups}

\begin{figure*}[t]
\centering
\includegraphics[width=1.55\columnwidth]{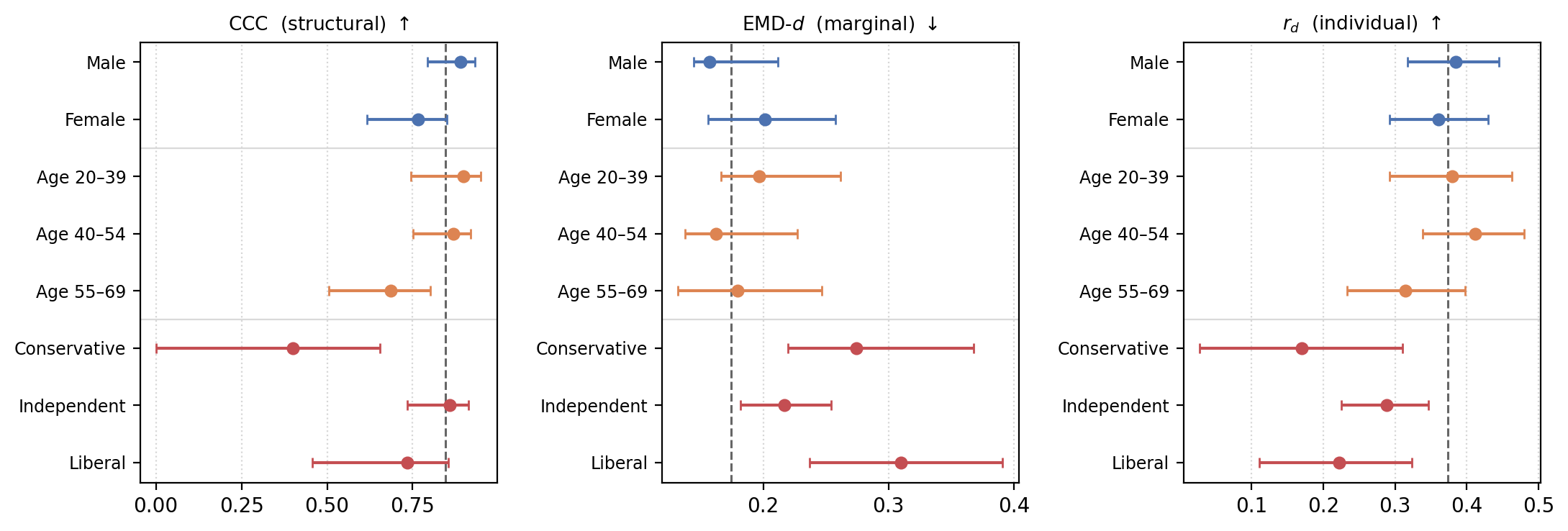}
\caption{Subgroup fidelity of LoRA + MLP across the structural (CCC), marginal (EMD-d), and individual ($r_d$) axes. Each point recomputes the fidelity axis by comparing the simulator against ground truth restricted to that same subsample (e.g. sim vs. GT among Conservatives only). Bars are 95\% bootstrap CIs; the dashed line marks the full-sample value.}
\label{fig:subgroups}
\end{figure*}

Aggregate fidelity can mask systematic failure on the very subpopulations a pluralistic simulator is meant to represent; a model can misrepresent minority groups \cite{wang2025large, li2025chatgpt}. We test this directly by recomputing all three axes within demographic subgroups for the best simulator (LoRA $+$ MLP), slicing along the subgroups based on gender, age, and political orientation (Fig.~\ref{fig:subgroups}). As the slicing variable is itself one of the 12 predictors, it is dropped from that slice's structural regression (11 predictors); all else is unchanged.

Recovery is \emph{not} uniform. Lin's CCC falls from $0.85$ overall to $0.40$ in the conservative subsample (and to $0.69$ in the oldest cohort), and the marginal/individual fidelity in the political-identity subsamples drops as well. Fidelity in several subgroups is markedly worse than the full-sample value (e.g.\ conservative CCC $0.40$ vs.\ $0.85$ overall), the signal that recovery differs across subgroups; small slices widen the CIs, so we read the gaps as \emph{suggestive}. The simulator thus appears to recover the \emph{aggregate} discernment distribution while recovering \emph{how} discernment is organized within some subgroups far less well, which is the risk the pluralistic alignment community warns against (full per-slice CIs in App.~\ref{app:full-cis}).

\section{Discussion}

\paragraph{LoRA $+$ MLP leads on the marginal axis.}
LoRA $+$ MLP attains the lowest cross-respondent EMD on all three per-respondent scalar summaries and is competitive on both individual fidelity metrics. But only the EMD gap separates cleanly; structurally and individually its CIs overlap ZS-perItem (Tab.~\ref{tab:full-cis}).

\paragraph{Output head appears to shape the simulation fidelity.}
LoRA and LoRA $+$ MLP share the same backbone and LoRA rank but differ only in output head, and in our single run that design choice coincides with a drastic difference in fidelity. The tentative lesson is not that fine-tuning fails or succeeds on its own, but that the output head appears to play a role.

\paragraph{PPI is conditional on bias variance and algebraically degenerate for fine-tuned simulators.}
PPI helps simulators that are far off but pulls already-calibrated simulators away from GT. PPI is beneficial mainly when the simulator's bias variance dominates the pilot's sampling variance. Beyond that empirical pattern, PPI has a structural problem with fine-tuned simulators that we make explicit: when $\hat y_{\text{pilot}} = y_{\text{pilot}}$ (the limit of training-set memorization), $\lambda_q = \mathrm{Cov}(y, \hat y)/\mathrm{Var}(\hat y) = 1$ identically and the rectifier reduces to $\hat\theta_q = \bar{\hat y}_{\text{held},q}$, the simulator's own held-out mean. The very thing fine-tuning does (drive pilot error to zero) defeats PPI's bias-estimation step.

\paragraph{Fidelity may not be pluralistically uniform.}
Subgroup-based evaluation (Sec.~\ref{sec:subgroups}) shows that recovery may not be uniform across subgroups, even though pooled fidelity is high, though small slices and wide CIs make this preliminary. A simulator can pass an aggregate audit while providing more distorted results on certain types of groups \cite{wang2025large, li2025chatgpt}, so fidelity per subgroup should be closely audited in future works.

\paragraph{Closing the individual-fidelity gap remains an open problem.} Across all methods, even the best simulator (LoRA $+$ MLP) reaches only $r_d{=}0.37$, accounting for $\sim$14\% of the cross-respondent variance in discernment. This suggests that, at least for this task, LLM silicon samples may not be suitable as direct substitutes for individual-level participant measurement. What level of individual fidelity is adequate and attainable for a given downstream use remains to be established by further empirical work.

\section*{Limitations}

\textbf{Single domain.} All experiments are on one survey (COVID-19 misinformation, South Korea, May 2020). Effects of cultural and linguistic context are unmeasured, and the levels of fidelity may differ in domains where the LLM's pretraining prior is more or less aligned. \textbf{Single backbone.} LoRA results use Qwen3-8B; closed-source frontier models with stronger zero-shot priors may narrow the prompting--fine-tuning gap. \textbf{Pilot composition.} Although we use a fixed-seed pilot, results might still be sensitive to the specific draw \cite{cummins2025threat}. \textbf{Unaccounted variance sources.} Our CIs resample respondents only, not the pilot draw, seed, or single fine-tuning run, so method rankings rest on single runs and should be read as indicative. \textbf{Recovery, not cognitive process.} High statistical recovery does not imply that LLM outputs reflect human cognitive processes or are appropriate for causal inference \cite{anthisposition, hwang2025human}; we recover statistical structure, not psychological mechanism.

\section*{Future Work}\label{sec:future}

The most promising next steps are: (i) a multi-domain replication; (ii) comparison among varying pilot sizes; (iii) alternative approaches such as Distribution Shift Alignment \cite{huang2025distribution} as a fine-tuning loss that targets distributional rather than per-cell objectives.

\section*{Impact Statement}

This paper presents work whose goal is to advance the understanding of large language models as instruments for social-science measurement. Our findings suggest that LLM-based survey simulations should not be treated as drop-in replacements for human respondents, and we provide methodology for evaluating their fidelity along multiple axes before they are used downstream. Misuse of synthetic survey data (treating uncalibrated LLM outputs as substitutes for human responses in policy decisions or intervention design) could amplify pre-existing biases or distort minority perspectives, as documented. By emphasizing recovery diagnostics across structural, marginal, and individual axes, our work aims to encourage more cautious and evaluation-driven use of LLM-based simulation in research practice.

\section*{AI tools usage disclosure}

During the preparation of this work, the authors used Claude for text refinement and code review. The authors reviewed and edited the content as needed and take full responsibility for the publication's content.

\bibliography{final}
\bibliographystyle{icml2026}

\newpage
\appendix
\onecolumn

\section{Example Prompt}\label{app:prompts}

The full participant block reproduces the seven psychometric / exposure construct items verbatim with item-text=label pairs. The 36 claims are presented it a per-respondent shuffled order. The per-item variant queries the same model 36 times per respondent, once per claim; the system message changes “exactly 36 labels” to “exactly 1 label” and the user message lists only one claim.

\begin{tcolorbox}[colback=gray!5, colframe=gray!50, title=Zero-Shot Batch Prompt, fonttitle=\bfseries\small]
\small
\textbf{System:} 

\begin{small}
\begin{verbatim}
You are a model that predicts a participant's perceived accuracy ratings for 
multiple claims based on participant information. 
The survey took place in South Korea in May 2020, during the COVID-19 pandemic.
Return strict JSON only with this schema: {"answers": ["<label>", "..."]} 
The "answers" array must contain exactly 36 labels in the exact same order as
the provided claims. 
Allowed labels: Not accurate at all, Not very accurate, Somewhat accurate, 
Very accurate, Have not seen it
Do not include explanations, reasons, claim IDs, or extra keys.

\end{verbatim}
\end{small}

\textbf{User:} 
\begin{small}
\begin{verbatim}
Claims to evaluate (in order):
- Korea's method of COVID19 diagnosis is inappropriate.
- COVID19 diagnostic test is free of charge for suspected patients.
- Hand-washing and social distancing is more effective in COVID19 prevention
  than wearing a mask.
- Foreign press including the BBC and the NYT reported that Korea is
  successfully coping with COVID19 through prompt diagnostic tests.
  ... (32 more claims) ...

Participant information:
Participant profile:
- Gender: Male
- Age: 61 years old
- Education: College (2-3 years)
- Household income: KRW 2M-3M
- Political orientation: Conservative

Pre-existing attitudes/perceptions:
- Open-mindedness: A person should always consider new possibilities=Slightly
  agree; People should always take into consideration evidence that goes
  against their beliefs=Slightly agree; ... (6 more items) ...
- Faith in intuition: I trust my gut to tell me what's true and what's
  not=Neither agree nor disagree; ... (3 more items) ...
- Need for evidence: ... (4 items, all Neither agree nor disagree) ...
- Truth as political construct: ... (4 items) ...
- Skepticism: I often accept other people's explanations without further
  thought=Slightly agree; It is easy for other people to convince me=Agree;
  ... (3 more items) ...
- COVID-19 information exposure: Daily newspapers=not at all;
  Television=very frequently; Online news=not at all; Social media=not at all;
  Health or medical professional websites=not at all; People around me
  (family, friends, coworkers)=frequently; Doctors=not at all
- COVID-19 emotional response: I feel fear about COVID-19=Quite a bit;
  I feel worried about COVID-19=Very much; I feel angry about COVID-19=Quite
  a bit; I feel hopeful about prevention and treatment of COVID-19=Very much
\end{verbatim}
\end{small}
\end{tcolorbox}

\newpage
\section{Bootstrap Confidence Intervals for All Body Tables}\label{app:full-cis}

For each headline metric we resample the per-method eval set with replacement, $n_{\text{boot}}{=}1{,}000$, fixed seed. On each resample we recompute the entire statistic from scratch -- per-respondent $d_i, m_i, t_i$ scalars; the 12 paired predictor--$d_i$ Pearson correlations and standardized OLS coefficients; CCC, sign-agreement, magnitude ratio; cross-respondent EMDs; per-respondent paired metrics ($r_d, \text{MAE}_d$). The reported point estimate is computed once on the full sample (not the bootstrap median); the CI bounds are the 2.5\% / 97.5\% percentiles of the bootstrap distribution.

The tables in the main paper report point estimates only for brevity. The same numbers with their 95\% percentile bootstrap CIs from $n_{\text{boot}}{=}1{,}000$ respondent resamples are tabulated below.

\subsection{Headline metrics (companion to Tab.~\ref{tab:headline})}
\begin{table}[h]
\centering
\small
\setlength{\tabcolsep}{4pt}
\begin{tabular}{l|cc|c|cc}
\toprule
 & \multicolumn{2}{c|}{\textbf{Structural}~$\uparrow$} & \textbf{Marginal}$\downarrow$ & \multicolumn{2}{c}{\textbf{Individual}}\\
\textbf{Method} & CCC$_{\text{biv}\,r}$ & CCC$_{\text{OLS}\,\beta}$ & EMD-$d$ & $r_d$ & MAE$_d$ \\
\midrule
ZS & 0.56 \textcolor{gray}{[0.43, 0.65]} & 0.50 \textcolor{gray}{[0.38, 0.58]} & 0.716 \textcolor{gray}{[0.683, 0.749]} & 0.12 \textcolor{gray}{[0.07, 0.16]} & 0.79 \textcolor{gray}{[0.76, 0.82]} \\
ZS-perItem & 0.80 \textcolor{gray}{[0.71, 0.85]} & 0.71 \textcolor{gray}{[0.58, 0.77]} & 0.484 \textcolor{gray}{[0.460, 0.505]} & 0.31 \textcolor{gray}{[0.25, 0.35]} & 0.58 \textcolor{gray}{[0.56, 0.60]} \\
FS & 0.51 \textcolor{gray}{[0.35, 0.65]} & 0.59 \textcolor{gray}{[0.45, 0.68]} & 0.226 \textcolor{gray}{[0.197, 0.256]} & 0.18 \textcolor{gray}{[0.13, 0.23]} & 0.67 \textcolor{gray}{[0.64, 0.69]} \\
FS-perItem & 0.61 \textcolor{gray}{[0.47, 0.72]} & 0.63 \textcolor{gray}{[0.51, 0.71]} & 0.412 \textcolor{gray}{[0.388, 0.438]} & 0.27 \textcolor{gray}{[0.22, 0.32]} & 0.58 \textcolor{gray}{[0.56, 0.61]} \\
LoRA & 0.60 \textcolor{gray}{[0.45, 0.70]} & 0.41 \textcolor{gray}{[0.24, 0.54]} & 0.633 \textcolor{gray}{[0.604, 0.664]} & 0.17 \textcolor{gray}{[0.11, 0.22]} & 0.75 \textcolor{gray}{[0.73, 0.78]} \\
LoRA + MLP & 0.85 \textcolor{gray}{[0.76, 0.89]} & 0.78 \textcolor{gray}{[0.68, 0.83]} & 0.174 \textcolor{gray}{[0.148, 0.216]} & 0.37 \textcolor{gray}{[0.33, 0.42]} & 0.61 \textcolor{gray}{[0.59, 0.64]} \\
\bottomrule
\end{tabular}
\caption{\textbf{Headline metrics with 95\% bootstrap CIs.} Same data as Tab.~\ref{tab:headline} but with bracketed gray 95\% percentile CIs from resampling held-out respondents.}
\label{tab:full-cis}
\end{table}

\subsection{Individual fidelity per subset}
\begin{table}[h]
\centering
\small
\setlength{\tabcolsep}{4pt}
\resizebox{\textwidth}{!}{%
\begin{tabular}{l|cc|cc|cc}
\toprule
 & \multicolumn{2}{c|}{\textbf{Discernment $d_i$}} & \multicolumn{2}{c|}{\textbf{Misinfo $m_i$}} & \multicolumn{2}{c}{\textbf{Trueinfo $t_i$}} \\
\textbf{Method} & $r\,\uparrow$ & MAE\,$\downarrow$ & $r\,\uparrow$ & MAE\,$\downarrow$ & $r\,\uparrow$ & MAE\,$\downarrow$ \\
\midrule
ZS & 0.12 \textcolor{gray}{[0.07, 0.16]} & 0.79 \textcolor{gray}{[0.76, 0.82]} & 0.14 \textcolor{gray}{[0.09, 0.19]} & 0.54 \textcolor{gray}{[0.52, 0.56]} & -0.01 \textcolor{gray}{[-0.05, 0.05]} & 0.40 \textcolor{gray}{[0.38, 0.41]} \\
ZS-perItem & 0.31 \textcolor{gray}{[0.25, 0.35]} & 0.58 \textcolor{gray}{[0.56, 0.60]} & 0.16 \textcolor{gray}{[0.11, 0.21]} & 0.45 \textcolor{gray}{[0.43, 0.47]} & 0.18 \textcolor{gray}{[0.11, 0.23]} & 0.32 \textcolor{gray}{[0.30, 0.33]} \\
FS & 0.18 \textcolor{gray}{[0.13, 0.23]} & 0.67 \textcolor{gray}{[0.64, 0.69]} & 0.16 \textcolor{gray}{[0.10, 0.21]} & 0.51 \textcolor{gray}{[0.49, 0.53]} & 0.04 \textcolor{gray}{[-0.01, 0.08]} & 0.38 \textcolor{gray}{[0.36, 0.39]} \\
FS-perItem & 0.27 \textcolor{gray}{[0.22, 0.32]} & 0.58 \textcolor{gray}{[0.56, 0.61]} & 0.20 \textcolor{gray}{[0.13, 0.25]} & 0.45 \textcolor{gray}{[0.43, 0.47]} & 0.11 \textcolor{gray}{[0.06, 0.16]} & 0.30 \textcolor{gray}{[0.29, 0.32]} \\
LoRA & 0.17 \textcolor{gray}{[0.11, 0.22]} & 0.75 \textcolor{gray}{[0.73, 0.78]} & 0.15 \textcolor{gray}{[0.09, 0.19]} & 0.58 \textcolor{gray}{[0.56, 0.60]} & 0.09 \textcolor{gray}{[0.04, 0.14]} & 0.31 \textcolor{gray}{[0.30, 0.33]} \\
LoRA + MLP & 0.37 \textcolor{gray}{[0.33, 0.42]} & 0.61 \textcolor{gray}{[0.59, 0.64]} & 0.34 \textcolor{gray}{[0.29, 0.39]} & 0.46 \textcolor{gray}{[0.44, 0.48]} & 0.15 \textcolor{gray}{[0.10, 0.20]} & 0.38 \textcolor{gray}{[0.37, 0.40]} \\
\bottomrule
\end{tabular}}
\caption{\textbf{Individual fidelity with 95\% bootstrap CIs.} For each per-respondent scalar (discernment $d_i$, Misinfo mean $m_i$, Trueinfo mean $t_i$), $r$ is the Pearson correlation between simulator and GT and MAE is the mean absolute deviation $|x_i^{\text{GT}}-x_i^{\text{sim}}|$.}
\label{tab:full-resp-cis}
\end{table}

\newpage

\subsection{Structural decomposition (companion to Tab.~\ref{tab:aux})}
\begin{table}[h]
\centering
\small
\setlength{\tabcolsep}{6pt}
\begin{tabular}{l|cc|cc}
\toprule
 & \multicolumn{2}{c|}{\textbf{Bivariate $r$}} & \multicolumn{2}{c}{\textbf{Standardized $\beta$}} \\
\textbf{Method} & sign-agree & $|$sim$|/|$GT$|$ & sign-agree & $|$sim$|/|$GT$|$ \\
\midrule
ZS & 0.58 \textcolor{gray}{[0.50, 0.83]} & 1.07 \textcolor{gray}{[0.92, 1.26]} & 0.58 \textcolor{gray}{[0.33, 0.75]} & 1.03 \textcolor{gray}{[0.88, 1.14]} \\
ZS-perItem & 0.83 \textcolor{gray}{[0.67, 1.00]} & 1.48 \textcolor{gray}{[1.31, 1.68]} & 0.75 \textcolor{gray}{[0.58, 0.92]} & 1.50 \textcolor{gray}{[1.28, 1.62]} \\
FS & 0.58 \textcolor{gray}{[0.33, 0.83]} & 0.56 \textcolor{gray}{[0.46, 0.73]} & 0.50 \textcolor{gray}{[0.33, 0.75]} & 0.63 \textcolor{gray}{[0.55, 0.83]} \\
FS-perItem & 0.92 \textcolor{gray}{[0.58, 1.00]} & 0.85 \textcolor{gray}{[0.73, 0.98]} & 0.67 \textcolor{gray}{[0.50, 0.92]} & 0.99 \textcolor{gray}{[0.84, 1.13]} \\
LoRA & 0.50 \textcolor{gray}{[0.42, 0.83]} & 0.50 \textcolor{gray}{[0.43, 0.67]} & 0.50 \textcolor{gray}{[0.33, 0.75]} & 0.58 \textcolor{gray}{[0.48, 0.75]} \\
LoRA + MLP & 0.92 \textcolor{gray}{[0.67, 1.00]} & 1.28 \textcolor{gray}{[1.14, 1.46]} & 0.75 \textcolor{gray}{[0.50, 0.92]} & 1.32 \textcolor{gray}{[1.12, 1.44]} \\
\bottomrule
\end{tabular}
\caption{\textbf{Structural-fidelity decomposition with 95\% bootstrap CIs.} Same data as Tab.~\ref{tab:aux} with respondent-level bootstrap intervals.}
\label{tab:full-struct-cis}
\end{table}

\subsection{PPI rectification (companion to Tab.~\ref{tab:ppi})}
\begin{table}[h]
\centering
\small
\setlength{\tabcolsep}{4pt}
\begin{tabular}{l|cc|cc}
\toprule
 & \multicolumn{2}{c|}{\textbf{Raw simulator}} & \multicolumn{2}{c}{\textbf{$+$ PPI rectification}}\\
\textbf{Method} & $\mu_\text{Mis}$ & $\mu_\text{Tru}$ & $\mu_\text{Mis}$ & $\mu_\text{Tru}$ \\
\midrule
\textit{GT} & 2.04 & 3.13 & 2.04 & 3.13 \\
\midrule
ZS & 1.77 & 3.45 & 2.12 \textcolor{gray}{[2.12, 2.12]} \textcolor{green!50!black}{$\downarrow$} & 3.08 \textcolor{gray}{[3.08, 3.08]} \textcolor{green!50!black}{$\downarrow$} \\
ZS-perItem & 2.01 & 2.98 & 2.12 \textcolor{gray}{[2.12, 2.12]} \textcolor{red}{$\uparrow$} & 3.08 \textcolor{gray}{[3.08, 3.08]} \textcolor{green!50!black}{$\downarrow$} \\
FS & 2.06 & 3.28 & 2.11 \textcolor{gray}{[2.10, 2.11]} \textcolor{red}{$\uparrow$} & 3.09 \textcolor{gray}{[3.09, 3.09]} \textcolor{green!50!black}{$\downarrow$} \\
FS-perItem & 2.02 & 3.13 & 2.11 \textcolor{gray}{[2.11, 2.11]} \textcolor{red}{$\uparrow$} & 3.07 \textcolor{gray}{[3.07, 3.07]} \textcolor{red}{$\uparrow$} \\
LoRA$^\dagger$ & 1.57 & 3.21 & 1.58 \textcolor{gray}{[1.57, 1.59]}  & 3.19 \textcolor{gray}{[3.18, 3.20]}  \\
LoRA + MLP$^\dagger$ & 2.01 & 3.23 & 2.03 \textcolor{gray}{[2.00, 2.05]}  & 3.21 \textcolor{gray}{[3.19, 3.22]}  \\
\bottomrule
\end{tabular}
\caption{\textbf{PPI rectification.} Same data as Tab.~\ref{tab:ppi} with respondent-level bootstrap intervals.}
\label{tab:ppi-cis}
\end{table}

\subsection{Subgroup analysis (companion to Fig.~\ref{fig:subgroups})}
\begin{table}[h]
\centering
\small
\setlength{\tabcolsep}{3pt}
\resizebox{.7\columnwidth}{!}{%
\begin{tabular}{lccccc}
\toprule
Subgroup & $n$ & EMD-$d$ $\downarrow$ & CCC$_{\text{biv}\,r}$ $\uparrow$ & $r_d$ $\uparrow$ & MAE$_d$ $\downarrow$ \\
\midrule
All respondents & 1392 & 0.17 \textcolor{gray}{[0.15, 0.22]} & 0.85 \textcolor{gray}{[0.76, 0.89]} & 0.37 \textcolor{gray}{[0.33, 0.42]} & 0.61 \textcolor{gray}{[0.59, 0.64]} \\
\midrule
\multicolumn{6}{l}{\emph{Gender}}\\
\quad Male & 719 & 0.16 \textcolor{gray}{[0.14, 0.21]} & 0.89 \textcolor{gray}{[0.79, 0.93]} & 0.38 \textcolor{gray}{[0.32, 0.45]} & 0.62 \textcolor{gray}{[0.59, 0.66]} \\
\quad Female & 673 & 0.20 \textcolor{gray}{[0.16, 0.26]} & 0.77 \textcolor{gray}{[0.62, 0.85]} & 0.36 \textcolor{gray}{[0.29, 0.43]} & 0.60 \textcolor{gray}{[0.57, 0.63]} \\
\addlinespace
\multicolumn{6}{l}{\emph{Age group}}\\
\quad Age 20--39 & 475 & 0.20 \textcolor{gray}{[0.17, 0.26]} & 0.90 \textcolor{gray}{[0.75, 0.95]} & 0.38 \textcolor{gray}{[0.29, 0.46]} & 0.63 \textcolor{gray}{[0.58, 0.67]} \\
\quad Age 40--54 & 483 & 0.16 \textcolor{gray}{[0.14, 0.23]} & 0.87 \textcolor{gray}{[0.75, 0.92]} & 0.41 \textcolor{gray}{[0.34, 0.48]} & 0.60 \textcolor{gray}{[0.56, 0.64]} \\
\quad Age 55--69 & 434 & 0.18 \textcolor{gray}{[0.13, 0.25]} & 0.69 \textcolor{gray}{[0.51, 0.80]} & 0.31 \textcolor{gray}{[0.23, 0.40]} & 0.61 \textcolor{gray}{[0.57, 0.66]} \\
\addlinespace
\multicolumn{6}{l}{\emph{Political orientation}}\\
\quad Conservative & 215 & 0.27 \textcolor{gray}{[0.22, 0.37]} & 0.40 \textcolor{gray}{[0.00, 0.65]} & 0.17 \textcolor{gray}{[0.03, 0.31]} & 0.61 \textcolor{gray}{[0.55, 0.68]} \\
\quad Independent & 812 & 0.22 \textcolor{gray}{[0.18, 0.25]} & 0.86 \textcolor{gray}{[0.74, 0.91]} & 0.29 \textcolor{gray}{[0.23, 0.35]} & 0.59 \textcolor{gray}{[0.56, 0.62]} \\
\quad Liberal & 365 & 0.31 \textcolor{gray}{[0.24, 0.39]} & 0.73 \textcolor{gray}{[0.46, 0.85]} & 0.22 \textcolor{gray}{[0.11, 0.32]} & 0.65 \textcolor{gray}{[0.60, 0.71]} \\
\bottomrule
\end{tabular}}
\caption{\textbf{Three-axis fidelity of the best simulator (LoRA $+$ MLP) within demographic subgroups} (95\% bootstrap CI). When the slicing variable is itself a structural predictor, it is dropped from the structural regression for that slice (11 predictors).}
\label{tab:subgroups}
\end{table}

\end{document}